\documentclass[oneside,english]{amsart}
\usepackage[T1]{fontenc}
\usepackage[utf8]{inputenc}
\usepackage{url}
\usepackage{amsthm}
\usepackage{graphicx}
\PassOptionsToPackage{normalem}{ulem}
\usepackage{ulem}

\makeatletter

\newcommand{\noun}[1]{\textsc{#1}}

\numberwithin{equation}{section}
\numberwithin{figure}{section}
\theoremstyle{plain}
\newtheorem{thm}{\protect\theoremname}
  \theoremstyle{definition}
  \newtheorem{example}[thm]{\protect\examplename}
  \theoremstyle{plain}
  \newtheorem{cor}[thm]{\protect\corollaryname}

\makeatother

\usepackage{babel}
  \providecommand{\corollaryname}{Corollary}
  \providecommand{\examplename}{Example}
\providecommand{\theoremname}{Theorem}

\begin{document}

\title{About $\tau$-chain}

\author{Ohad Asor}
\begin{abstract}
$\tau$-chain (pronounced tau-chain) is a decentralized peer-to-peer
network having three unified faces: Rules, Proofs, and Computer Programs,
allowing a generalization of virtually any centralized or decentralized
P2P network, together with many new abilities, as we present on this
note.\end{abstract}
\maketitle
\begin{quotation}
\emph{“If law-making is a game, then it is a game in which changing
the rules is a move.” - Peter Suber presenting Nomic \cite{key-12}.}
\end{quotation}
Document version: 0.2 Feb 6 2015

Document version: 0.21 Feb 9 2015

Document version: 0.22 Feb 12 2015

\section{Introduction }

We %
\footnote{My Logic teacher HunterMinerCrafter \url{https://bitcointalk.org/index.php?action=profile;u=245263},
who introduced this idea and design to me, and myself, Ohad Asor.%
} propose a network that unifies computer-readable knowledge, rules,
inference, reasoning, and more equivalent aspects. The network evolves
with time toward gathering more meaningfull information.

\subsection{What and Why Do We Need to Decentralize?}

Humanity is not wild, but engineered. It is, and always was, engineered
and manipulated by relatively small groups of people. Our media, education,
economy, law, politics, ethics, are all shaped mainly according to
the views of small groups. Those groups tell us we are englightened,
but in fact, the situation is very much of opposite nature. Immanuel
Kant began his famous essay ``What is Enlightment?'', and he defines:
\begin{quotation}
\emph{``Enlightenment is man's emergence from his self-imposed nonage.
Nonage is the inability to use one's own understanding without another's
guidance. This nonage is self-imposed if its cause lies not in lack
of understanding but in indecision and lack of courage to use one's
own mind without another's guidance. Dare to know! (Sapere aude.)
\textquotedbl{}Have the courage to use your own understanding,\textquotedbl{}
is therefore the motto of the enlightenment.''}
\end{quotation}
Those 1784 (1984?) words are in fact relevant for today more than
ever: today we can break the chains of opression and let our own voice
emerge.

About a century later, Franz Kafka described the inability of the
individual to stand against irrational and moraless bureaucratic systems.
While the laws are never actually given to the citizens, they still
must obey them or otherwise suffer the wrath of society acting in
violence against them, while the system contradicts itself: on one
hand they claim they can never define law in a closed form, and on
the other hand, they always find a way to justify their actions according
to the law. Today, they also blame their computers and the records
or decisions it allows to keep or modify. So maybe law cannot be formalized
for us, but can be formalized for them?

Law can and should be formalized. The most important property a law
system should have is a consistent ethical basis (e.g. constitution)
and consistent implications from this basis to laws themselves. But
who should formalize the laws?

Formalizing laws will pose even a greater danger if done by centralized
hands. It is evident that democracy is incapable of assisting: the
way in which voting is done today is far from assuring a consistent,
moral, and functioning system. 

Until recent times, this was indeed an inevitable situation. Now we
have mathematical and technological ways to create laws collaboratively,
while preserving on frames we set for ourselves, like consistency,
votes, or minimal requirements.

Centralization of law is only one thing. What about centralization
of information? Think of the following situation: Google has the most
valuable information in the world, namely, ``what people want''.
They do not give us access to their databases. But imagine, what if
we could access their data: we would not only obtain this important
information, but also have the ability to perform a much more sophisticated
search. We could, for example, query about topics related to a given
topic, or automatically build new aggregated data from the database,
and basically have endless additional uses.

Communication is also centralized. When communicating by electronic
means, what we say is often intercepted by unintended parties. Our
privacy is deeply vulnerable. We do not tend to have our own website
and post our thoughts there, but we do it on centralized locations
like Facebook. We do it because centralized hands give us better technology,
with long term support and less bugs. We pay for it with our privacy,
endless ads and marketing junk, and we even let them manipulate who
our friends will be. Can we keep all benefits and have a moral high-end
software?

\subsection{The Vision}

$\tau$-chain's goal is to unite, yet keeping decentralization, humanity's
knowledge and thinking, know-hows and communication, laws and opinions,
all into one giant shared database that is able to be coherent and
consistent, to be queried meaningfully, to reuse information/code/data
efficiently, to allow all kind of social operations and communications
to be done with no unwanted guests, to allow every user set their
own specific rules and to be part of communities sharing the same
thoughts, or goals, or needs. $\tau$-chain does not enforce anyone
to subscribe or follow one or another ruleset. But it lets all mental
and technological benefits to be combined for the sake of better humanity.

Obviously, not all questions are solvable, some better solved by humans
and some better solved by machines, and some are solvable with many
machines and a lot of knowledge. Once one is able to solve a puzzle,
namely: to prove a given claim, it gets into the network. Until then,
it stays out.

This vision is not new, and probably began with the Semantic Web as
we will describe on section 5, but could not be fulfilled without
the nature of decentralization, which also gives the ability to fairly
incentivize participants. Moreover, this vision in its decentralized
version is not new as well, but no serious attempts were made before
to make it happen.

It is about time to begin doing to many issues bothering humanity
what Satoshi Nakamoto began to do to the monetary system.

It should be noted that $\tau$-chain isn't a coin as for itself,
and will bootstrap as a decentralized network, where developers are
just like everyone, without any kind of so-called premine or any currency
in hand, since the system itself has to be bootstrapped practically
without rules.

\section{Abbreviations and Definitions}
\begin{itemize}
\item db: Database 
\item P2P: Peer-to-Peer
\item prop{[}s{]}: Proposition{[}s{]} - a claim or definition to be understood
under strict logic %
\footnote{\textquotedbl{}Cats are either flowers or mammals\textquotedbl{} is
a true prop under pure logical interpretation.%
}
\item DHT: Distributed Hash Tables
\item DTLC: Dependently Typed Lambda Calculus
\item TFPL: Totally Functional Programming Language
\item RDF: Resource Description Framework
\item N3: Notation3 Language
\end{itemize}

\section{Overview}

We present $\tau$-chain, a fully decentralized P2P network being
a generalization of many centralized and decentralized P2P networks,
including the Blockchain. We aim to generalize the concept as much
as we find, and give users the ability to implement virtually any
P2P network over $\tau$-chain. Its interpretations, uses, and consequences
are far from being a P2P network only, and include software development,
legal, gaming, mathematics and sciences, logic, crypto-economies,
social networks, rule-making, democracy and votes, software repositories
(like decentralized Github+Appstore/Google Play), decentralized storage,
software approval and verification, even ``doing your homework in
History or Math'' in some sense (stronger sense that search engines),
and many more aspects %
\footnote{Zennet Supercompter's design has now changed to work over $\tau$-chain.
BitAgoras will work over it as well.%
} .

\subsection{Five Equivalent Definitions }

$\tau$-chain can be defined by several equivalent definitions: 
\begin{itemize}
\item A shared db of rules, with a client that is able to change the rules,
obey the rules, infer new rules from given rules, and make sure rules
are consistent. 
\item A shared db of computer programs' code being collaboratively composed
with revision control and custom permissions (like \noun{git}), with
a client that is able to run code, reuse code from existing programs,
and verify programs against formal specifications which are programs
themselves. 
\item A shared db of props, with a client that is able to state new props,
prove props with custom derivations rules which are props themselves,
and verify proofs to given props.
\item A shared db of ontologies, which are definitions of types (taxonomies)
and their relations, with a client that is able to propose new ontologies,
make sure that ontologies are consistent, and query the ontologies
db various queries.
\item A decentralized Nomic game.
\end{itemize}

\subsection{The Essence of $\tau$-chain node}

\begin{figure}[h]
\caption{$\tau$-chains}
\includegraphics[scale=0.2]{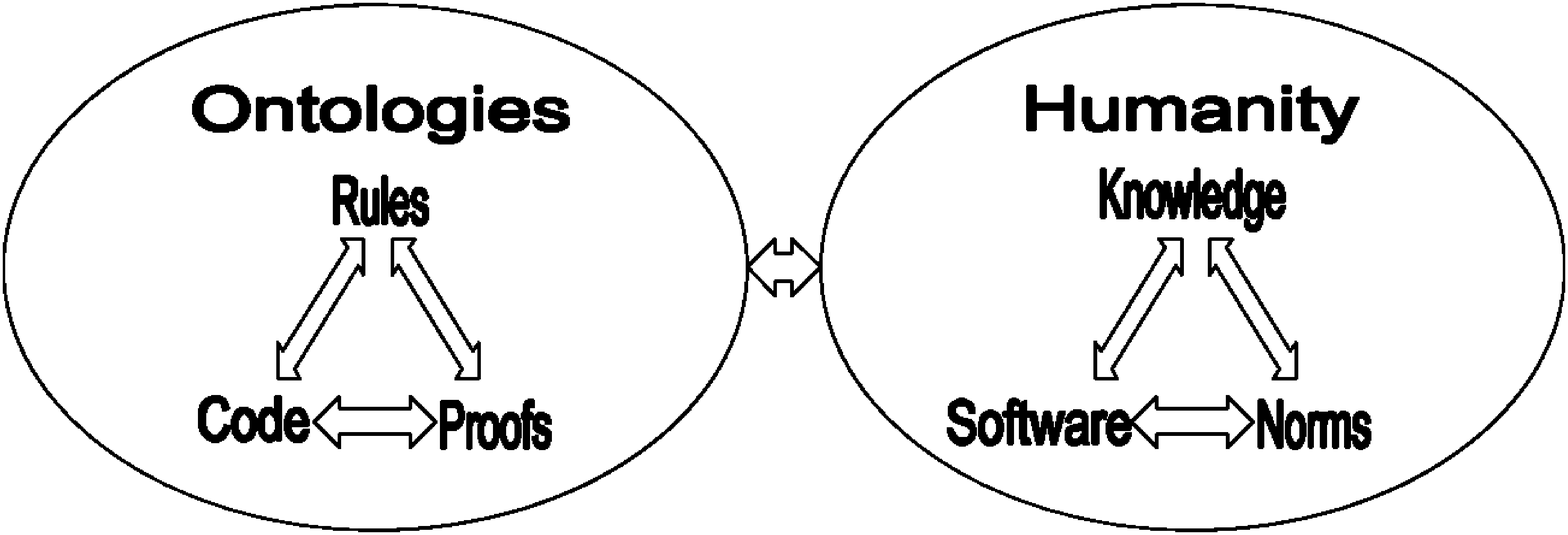}
\end{figure}
 At its basis, $\tau$-chain is nothing but a db of quads being tuples
of four words each, namely: context, subject, predicate, object. On
this note we show how rules, programs, proofs, and ontologies are
all naturally representable by a unified way, and point to some of
the far-reaching uses of such a system. 

Every node contains knowledge, at the very intelligent sense, namely
the ability to make inference and implication using logic. They do
not aim to represent a human intelligence, but their knowledge is
first and foremost how to communicate and transfar liable information.
Technically, this is can be done by formalizing with ontologies algorithms
like DHT and Blockchain. But when looking at this in a unfied way
as virtual creatures being able to ask rather retrieve, know rather
store, talk rather comminucate, consider rather verify, having only
mouth and ears and ontologies and a reasoner.

\section{Basic Equivalence Relations}

We give an informal explanation of rules as programs as proofs. For
a formal derivations, we refer the reader to the literature of Type
Theory.

By intuition, \textquotedbl{}running a program\textquotedbl{} can
easily be seen as equivalent to \textquotedbl{}obeying the rules\textquotedbl{},
while the rules are the code itself. The non-triviality lies on the
other direction: can all rules be formalized as programs? The answer
is positive, in a sense, and we will explain it through the proofs
interpretation.

\subsection{What are Proofs?}

From abstract logic point of view, a proof is a path taken from given
hypotheses and axioms, and ends up with a result to be proved under
the hypotheses, while the path has to be taken according to given
derivation rules, and follows the notion of \uline{implication}
($a\implies b$).
\begin{example}
Following are examples of derivation rules: \end{example}
\begin{itemize}
\item Modus Ponens: if A implies B and A is true, then B is true: 
\[
\left(\left(A\implies B\right)\wedge A\right)\implies B
\]

\item Cryptographic signature: it is impossible to generate a valid signature
(up to assumed validation procedure) without access to the corresponding
private keys. 
\item A mapping from data to its hash is one-to-one and noninvertible. Obviously,
this is false under pure mathematical logic, but it is valid in cryptography
being a practical science.
\item Hilbert's formal inference rules and Gentzen's formal inference rules.
\end{itemize}
It should be noted that derivation rules can be stated as axioms,
and axioms as hypotheses. Hence, a proof is inferring a statement
from other statements, or equivalently, deriving a rule given other
rules. Looking at definitions (taxonomy) as rules themselves, we see
that rules are proofs and proofs are rules, and that proving is inferring.

\subsection{Proofs=Programs}

This non-trivial result is the celebrated Curry-Howard isomorphism
. As there are many kinds of programs and many kinds of proofs, we
are interested in a space where programs always halt. Roughly speaking,
such programming language fall under the class of TFPL, e.g. \noun{Idris},
\noun{AGDA}, \noun{coq} %
\footnote{Though COQ isn't really DTLC but Calculus of Constructions, yet those
differenced doesn't really matter for our sake%
}, which correspond to DTLC. Such and other class are isomorphic to
the classes of proofs. 

TFPL are not Turing complete. This lets us escape from many paradoxes
arising from Turing machines formalism, promises us that programs
always halt, and give us strong abilities to claim and prove claims
about the program %
\footnote{Sometimes like ``the program is doing X and not doing Y in Z steps''.%
} . By this, we can prove that a given code satisfies a given unit-test
\footnote{Being a form of formal specification.%
} . We can also prove the execution of the program %
\footnote{cf. \foreignlanguage{english}{\url{http://amiller.github.io/lambda-auth/}}\selectlanguage{english}%
}. 

It should be noted that practiaclly, not being Turing complete at
this sense gives only advantages: any application one can come up
with does not require Turing completeness, but DTLC is enough. Turing
completene languages can do things that DTLC cannot only at extreme
(mostly infinite) theoretical cases.
\begin{cor}
Logical proposition can be interpreted as rules, and vice versa. Proofs
from axioms can be interpreted as inferring rules from other rules,
and vice versa. Computer programs written in a totally functional
programming language can be interpreted as constructive proofs, and
vice versa.
\end{cor}
\begin{figure}
\caption{Obeying Code}

~

\includegraphics[clip,scale=0.2]{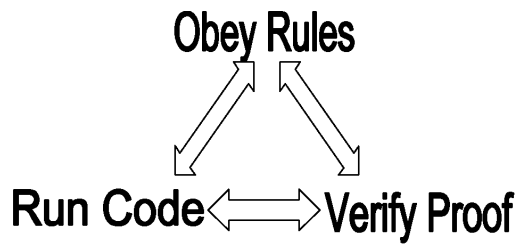}
\end{figure}

It should be stated that the correspondence between proofs and computer
programs is so strong, that it gives rise to new foundation of mathematics,
rather Cantor's/ZFC Set theory, namely: Category theory. It turns
out that proofs and programs are isomorphic to category theory, at
some sense. This is the Curry-Howard-Lambek isomorphism, aka Computational
Trinitarianism %
\footnote{\url{http://ncatlab.org/nlab/show/computational+trinitarianism}%
} .

\section{Ontologies and the Semantic Web}

\subsection{Background}

This is really a long story, and we're going to make it very relatively
short. It all began when the inventor of the World Wide Web, Tim Berners-Lee,
came up with a vision that all data online will be machine readable,
and build a web of relations between objects and types. That's the
Semantic Web vision in a nutshell. For this goal, a vast amount of
tools has been developed and still being developed. 

Nowadays there's a huge and growing amount of ontologies from all
aspects of life: basic web components and logical entities and relations
(OWL), legal (cf. LKIF ontologies), software, medicine, e-commerce,
cyptography, social media, geography, news, and many more %
\footnote{\url{http://www.w3.org/standards/semanticweb/ontology} \url{http://www.w3.org/wiki/Ontology_Dowsing#Lists_of_ontologies_and_services}%
}%
\footnote{Ontologies and search engines for ontologies from EulerGUI manual
\url{http://eulergui.sourceforge.net/documentation.html##Finding}%
}. Another notable project is \emph{dbpedia}, that already formalized
millions of concepts from Wikipedia with consistent ontologies. There
is also a vast amount of tools to manipulate RDF data, a list can
be found on the Semantic Web Wiki%
\footnote{\url{http://semanticweb.org}%
} .

The same growth can be observed in mathematics and other sciences
formalized for \noun{coq}, all naturally translatable to RDF, and
containing thoushands of mathematical theorems with their full proofs.

\subsection{Semantic Web's Semantics}

The basic representation of concepts is standardized by RDF (Resource
Description Framework). RDF is a language expressing Ontologies. An
ontology contains definitions of types (taxonomy) like ``dog is an
animal'', and relations between them like ``all dogs has four legs''.
Of course, a definition of Four has to make sense as well, like the
definition of ``has''.

Notation3 (N3) is a language makes it more convenient and human-readable
to write RDF ontologies. It is logically powerful enough to represent
DTLC. Moreover, it can always be converted into quads and vice versa,
which is the natural format for $\tau$-chains, as mentioned at the
end of the introduction.

RDF being exactly pure logic can also be converted into English, in
some limited sense, and vice versa. It is more machine-readable english
than human-readable, but still very natural. We of course speak about
the Attempto project.

\subsection{State-of-the-Art}

EulerGUI is a veteran IDE for various reasoning engines, demonstrating
the power of ontologies. Some of its supported formats are RDF, RDFS,
N3, OWL, UML, SPARQL, Attempto (Controlled English), and even Java
jars. It can output Java code that builds a UI according to an ontology,
using Rule based SWING. It supports four different and powerful reasoning
engines, logican queries, explained proofs, consistency checks, fuzzy
logic, graphical visualization, and more. It can be integrated with
the Deductions engine %
\footnote{\url{http://deductions.sourceforge.net}%
} which is itself written in ontologies. It is described on their website
as ``Artificial Intelligence techniques applied to common software
tasks, using First Order Logic through N3 + OWL ontologies and rules''.
Both EulerGUI and Deductions are compatible with the powerful reasoning
engine Drools. On Drools website one can find pros and cons of ontologies
based development %
\footnote{\url{http://docs.jboss.org/drools/release/5.4.0.Final/drools-expert-docs/html/ch01.html##d0e384}%
} . The reader is invited to get introduced to this world from the
linked materials. A more recent tool is Protege, a professional open-source
ontologies IDE.

\subsection{Ontologies of Rules}

Given we formalize our rules as an ontology. What are we going to
do with them? We take an example tool called cwm%
\footnote{\url{http://www.w3.org/2000/10/swap/doc/cwm.html}%
}. We can ask it questions that are answerable from within the rules,
even if not explicitly but by inference. We can also verify the consistency
of rules %
\footnote{An important example for consistency check is, given a logical proof
of some theorem represented as rules, verifying the proof is equivalent
to veriying the rule's consistency.%
} .

The reader is invited to take a look at its various tools%
\footnote{\url{http://attempto.ifi.uzh.ch/site/tools}%
}, including the reasoner and the OWL Verbalizer (OWL is RDF with some
basic ontologies defined, and is a W3C standard). See appendix for
demonstrative screenshots.

\subsection{Programs as Ontologies}

DTLC based languages can implement almost everything, and in practical
sense - everything. While Turing Complete languages suffer from undecidability,
on DTLC languages we can look at the code as logic we can work with
and prove various useful claims directly from the code, like that
the program does not use the internet, or accesses only certain files,
, show it halts, prove its execution path etc.

The transition from DTLC to RDF isn't trivial but two examples of
how it can be done can be found at \cite{key-4-1,key-2}. The structure
of the program (almost) doesn't change and namings can be preserved.
The RDF format can be kept for being human readable, but for machine-proofs
we do not stay on the RDF representation but use SMT solvers for reasoning.

\section{Peer-to-Peer}

As stated above, $\tau$-chain can generalize any P2P network. A decentralized
P2P client is a state machine that decides what to do given various
inputs, according to some rules.
\begin{example}
DHT is an example of a P2P architecture. One of the products using
it is BitTorrent. The most common flavor of DHT is Kademlia. Bitcoin's
network is DHT itself, but with full replication per node, while on
regular DHT the number of replicas can be controlled. A sample formal
specification (namely: rules) of the Kademlia can be found at \cite{key-14}
. Note that this specification is given with a TFPL, hence can easily
be translated into ontologies. We can see it has four primitives:
\noun{Ping}, \noun{Store}, \noun{FindNode}, and \noun{FindValue}.
The rules define what to do on each case.
\end{example}
~
\begin{example}
Bitcoin is a decentralized P2P network having proofs as its main interpretation,
while of course can be written as rules.
\end{example}
On $\tau$-chain, every node stores three local ontologies: its routing
table, its user's input, and the input from its peers. Those are combined
with the ontologies this node subscribes to, from the shared db, together
inferring what the client should do now, which can be local storage
operations or sending information to peer(s). It is like: I tell you
part of what I know, you think about it and tell me the conclusions
you want to share with me.
\begin{cor}
Any P2P network is a ruleset defining what to send given what was
received.
\end{cor}
A P2P network can set its own rules, and many contexts for many rules,
and by this on-going implementing new networks over this one, while
the user can decide on which contexts they want to participate. This
is $\tau$-chain.

\section{$\tau$-chain}

$\tau$-chain has two stages: bootstrapping and maturity.

\subsection{Bootstrapping}

The network will begin with the following ``bare'' client: it will
contain an ontology that implements simple DHT, and will run actions
by querying the ontology, using \noun{cwm}, queries like ``what should
I do now'', or more precisely, ``what should I send to each of my
peers'' and ``what should I do with my local storage''. From this
point we can implement everything by ontologies, in a collaborative
work, over a decentralized network.

On this bootstrapping stage we plan to insert many readily-made and
ontologies, and we plan to gather a round-table of professionals to
set this network's global rules together. Here are some possible features
to be implemented at the bootstrapping step:
\begin{itemize}
\item Rules to avoid malicious use of the network.
\item Ability to create separate contexts, where each context has its own
rules and they do not interfere. By this, users may create many programs
and users can pick which programs they want to use. Like decentralized
Appstore/Google Play.
\item Incentivizing every node for its work.
\item Implementing Bitcoin as a proof of concept%
\footnote{Nothing intended to be used in real life. Of course, improved cryptocurrency
is planned as further steps, especially at the scope of Agoras.%
}.
\item Distributed large-scale storage, which is essential for the system
itself, since it is planned to deal with large amount of data.
\item Voting.
\item How rules are going to be set from now on.
\end{itemize}

\subsection{Maturity}

Once the system is thought to have enough strength to go fully public
and for everyday use, we can speak of what can be done over it then.
\begin{itemize}
\item Decentralized Source Repository (e.g. decentralized \noun{github}).
\item Decentralized Application Repository.
\item Safely and automatically offer coins %
\footnote{$\tau$-chain will not implement coins from day 1, but it is something
that can and should be built upon the system.%
} to a human or machine proving a theorem or writing a software given
formal specifications.
\item Huge db of code fragments ready to be automatically reused by verifying
formal requirements or by detecting isomorphisms %
\footnote{Those isomorphisms can connect ideas across science. It is possible
that one writing some code that walks on graphs, will solve a problem
that a biologist is dealing with.%
} between ontologies.
\item Any kind of votes for any purpose, like development team vote for
a valid and authentic release, or decentralized democracy.
\item Collaborative social/corporate rule-making, with the ability to find
contradictions, to ask ``what is missing in order to obtain X''.
And voting for those rules if wanted.
\item Fully-customizable, safe, decentralized and private social networks,
as well as private clubs (even with membership fees, entrance test,
acceptance rules etc.).
\item Ask human-readable questions about virtually anything. Like ``where
did Aristotles live''. \noun{cwm} is able to answer such questions
from ontologies, and specify a proof for its answer. Of course, other
tools are both compatible and much more powerful, e.g. \noun{coq}.
\end{itemize}

\section{Conclusions}

We have shown how $\tau$-chain is able to generalize any collaborative
work, especially peer-to-peer networks. It provides ultimate information
sharing capabilities, with rich ways to query data, infer new information,
and act collaboratively. Existing P2P networks can be ported into
it, e.g. Bitcoin's Blockchain, and make them being controlled by additional
rules that can be changed on-the-fly with any rule for changing the
rules. It may serve as a universal source of trustable information,
as a collaborative source of knowledge, source code, and rules, in
a form that is both machine and human readable and processable. It
offers a ground in which sciences can be unified, and, more importantly,
people's thoughts can be met and unified cleanly.

\end{document}